\pdfoutput=1

\documentclass[11pt]{article}

\usepackage{EMNLP2023}

\usepackage{times}
\usepackage{latexsym}
\usepackage{graphicx}

\usepackage{booktabs}

\usepackage[T1]{fontenc}

\usepackage[utf8]{inputenc}

\usepackage{microtype}

\usepackage{inconsolata}

\newcommand{\aoc}{\textsc{ao-childes}}

\title{Evaluating Neural Language Models as\\ Cognitive Models of Language Acquisition}

\author{H\'{e}ctor Javier V\'{a}zquez Mart\'{\i}nez$^1$~~~Annika Heuser$^1$~~~Charles Yang$^{1,2}$~~~Jordan Kodner$^{3}$ \\
$^1$University of Pennsylvania, 
  Deptartment of Linguistics\\
$^2$University of Pennsylvania, 
  Deptartment of Computer and Information Science\\
  $^3$Stony Brook University, 
  Dept. of Linguistics \& Inst. for Advanced Computational Science\\
\texttt{hjvm@sas.upenn.edu}~~~\texttt{aheuser@sas.upenn.edu}\\
\texttt{charles.yang@ling.upenn.edu}~~~\texttt{jordan.kodner@stonybrook.edu}}

\begin{document}
\maketitle
\begin{abstract}
The success of neural language models (LMs) on many technological tasks has brought about their potential relevance as scientific theories of language  despite some clear differences between LM training and child language acquisition. In this paper we argue that some of the most prominent benchmarks for evaluating the syntactic capacities of LMs may not be sufficiently rigorous. In particular, we show that the template-based benchmarks lack the structural diversity commonly found  in the theoretical and psychological studies of language.  When trained on small-scale data modeling child language acquisition, the LMs can be readily matched by simple baseline models. We advocate for the use of the readily available, carefully curated datasets that have been evaluated for gradient acceptability by large pools of native speakers and are designed to probe the structural basis of grammar specifically. On one such dataset, the LI-Adger dataset, LMs evaluate sentences in a way inconsistent with human language users. We conclude with suggestions for better connecting LMs with the empirical study of child language acquisition. 
\end{abstract}

\section{Introduction}

The growth of neural language models (LMs) for NLP over the past decade has been followed by a growth in research on the potential of these  models  to provide insights into the cognitive aspects of human language acquisition, representation, and processing \citep{linzen2021syntactic}. Good, even human-like, performance on NLP tasks does not necessarily imply that LMs solve these in human-like ways, so computational linguists have designed a wide variety of experimental paradigms to probe specific properties of the models' linguistic knowledge \citep{linzen-etal-2016-assessing,chowdhury-zamparelli-2018-rnn,gulordava-etal-2018-colorless,wilcox-etal-2018-rnn,mccoy-etal-2020-syntax,hu-etal-2020-systematic,warstadt-etal-2020-blimp-benchmark,papadimitriou-etal-2021-deep,huebner-etal-2021-babyberta} These range from ways of classifying or extracting structures from internal representations \citep[e.g.,][]{hewitt-manning-2019-structural, tenney-etal-2019-bert,tucker-etal-2021-modified,papadimitriou-etal-2021-deep}, to building tasks inspired by psycholinguistic processing studies and classic acceptability rating task that theoretical linguists use to infer grammatical knowledge \citep[e.g.,][]{linzen-etal-2016-assessing,warstadt-etal-2020-blimp-benchmark,huebner-etal-2021-babyberta,sinclair-etal-2022-structural}. 

\begin{figure}[t]
    \centering
    \includegraphics[width=7.75cm]{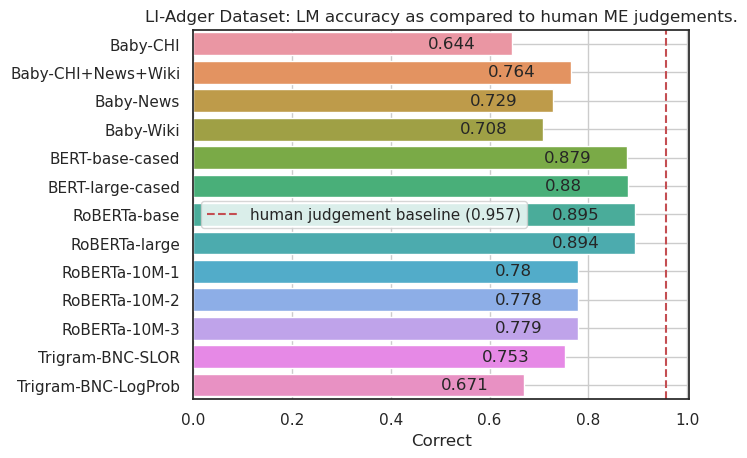}
    \vspace{-0.7cm}
    \caption{LM performance on the LI-Adger dataset. Human performance is marked by the vertical line.  Baby=BabyBERTa, CHI=\aoc, News=\textsc{ao-newsela}, Wiki=Wikipedia-1.}
    \label{fig:li-adger_performance}
\end{figure}

Of these approaches, acceptability rating may be the most popular. Large acceptability rating data sets focusing on syntax, semantics, and morphology, such as BLiMP \citep{warstadt-etal-2020-blimp-benchmark}, SyntaxGym \citep{gauthier-etal-2020-syntaxgym}, and CoLA \citep{warstadt-etal-2019-neural} lend themselves to benchmarking, and these sit alongside myriad smaller scale studies focused on specific lingusitic phenomena \citep[e.g.,][]{linzen2016assessing,marvin-linzen-2018-targeted,wilcox-etal-2018-rnn}. Results have been impressive for the most part. It appears, from the logic of these studies, that many state-of-the-art neural models are capable of inducing human-like grammatical knowledge on unannotated data\,--\,like  children during language acquisition. 

\subsection{Implications for Language Acquisition?}

Neural model training differs from human language acquisition in key ways, perhaps most obviously, in that most models are trained on orders of magnitude more input (in plain text form) than humans receive (in spoken or signed form)\--- BERT was trained on about 3.3B forms, and Chinchilla on 1.4T, while an English-learning child only receives about 10M word per year, for a total vocabulary measured in the hundreds at age three \citep{Fenson1994,Bornstein2004}.

Recent studies have begun to address this. Can we build models that learn from input on the scale of language acquisition? Would these models then inform our understanding of human language acquisition? \citet{warstadt2022artificial} favor this perspective. They argue that a computational model that performs well on behavioral probing benchmarks when trained on ablated input, that is at least as limited as a human learner's input, is evidence that the model is a good proxy for human linguistic knowledge. \citet{huebner-etal-2021-babyberta} showed that a specially tuned model trained on only 5M tokens of child-directed speech (CDS) performs well on a purpose-designed data set. And in 2023, an aptly-named shared task, the CoNLL/CMCL BabyLM Challenge,\footnote{\url{https://babylm.github.io/}} is asking participants to train on only 100M words (about the input of an adolescent) before testing on acceptability benchmarks. 

\subsection{Goals of the Paper}

A push towards extracting performance on smaller training data is a welcome change for the field. In addition to its possible cognitive implications, the drive will also benefit efficient NLP and NLP for low-resource languages. However, while we look forward to the impending engineering advances, we also urge caution in the approaches used to draw scientific conclusions about the nature of neural models' linguistic knowledge. In particular, we take issue with \citet{warstadt2022artificial}'s assertion that 
``positive results from model learners are more meaningful than negative results.''

Their reasoning follows that of an existence proof. If a model that strictly lacks any advantages over humans nevertheless succeeds at a task that requires human-like linguistic knowledge, then it is proof that there exists at least one model with human-like linguistic knowledge. A failure only tells us that this model failed for some reason that may or may not be relevant to the question at hand. 

However, this line of reasoning requires faith in the evaluation. If there are any potentially unrecognized non-human-like ways to succeed at the task, or if the task does not truly reflect acquisition, or the task does not actually test a relevant structural property of language, then a positive result becomes inconclusive at best. Unexpected shortcuts emerging from unforeseen biases in evaluation abound across NLP \citep{chao-etal-2018-negative,mccoy-etal-2019-right,wang-etal-2022-identifying}, so this is a realistic concern. Even the underlying reasoning that ``if a (neural) model X behaves like cognitive system Y, then it is equivalent to Y'' may be fraught \citep{guest2023logical}.

In this paper,\footnote{Our evaluation code and data are available at \url{https://github.com/hjvm/benchmarking_acquisition.git}} we evaluate LMs as models of language acquisition on two benchmarking data sets: the widely used Benchmark of Linguistic Minimal Pairs (BLiMP; \citealp{warstadt-etal-2020-blimp-benchmark}), which also forms part of the evaluation for the BabyLM Challenge, and Zorro \citep{huebner-etal-2021-babyberta}, a data set inspired by BLiMP with restricted vocabulary for acquisition-inspired models trained only on CDS.

Section \ref{sec:knowledge} reviews the nature of linguistic knowledge and child language acquisition. In Section \ref{sec:linear}, we introduce the BLiMP and Zorro benchmarks and subject them to baseline tests by simple non-human-like models. These establish several weaknesses in the organization and content of both benchmarks.  In Section \ref{sec:variability}, we evaluate neural models on a more challenging data set derived directly from theoretical linguistics papers. We find that LMs are not necessarily human-like in terms of within- and across-model variability. Finally, Section \ref{sec:discussion} concludes with a discussion of the logical problem of behavioral probing. We argue for \textbf{(a)} benchmarks that better probe the structural knowledge of syntax, \textbf{(b)} tests that reflect the developmental findings of language acquisition, and \textbf{(c)} more baseline models.

\section{Knowledge of Language and its Acquisition}\label{sec:knowledge}
One of the goals of linguistic theory is to characterize the properties that distinguish grammatical from ungrammatical sentences in a language. The empirical study of grammaticality, however, mainly relies on native speakers' acceptability judgments, which interact with other cognitive and perceptual systems and generally produce gradient results. For example, longer and more complex sentences, even when fully grammatical, are rated as less acceptable than shorter and simpler sentences. Nevertheless, large-scale investigations have established the structural basis of a categorical grammar \cite{Sprouse2013}. For example, syntactic constraints that prohibit certain transformational processes are shown to have a ``super-additive'' effect that go beyond acceptability rating due to sentence length and other non-structural factors.  Furthermore, acceptability judgments collected at scale are highly consistent with the data reported in the theoretical literature typically gathered informally with few consultants \citep{sprouse2012assessing,sprouse2013comparison,sprouse2017design}.

The structural basis of language and its uniformity across the linguistic community can be better appreciated from the perspective of child language acquisition.  Recent years have seen renewed interest in individual differences across child learners \cite{Kidd2018}, especially with respect to vocabulary acquisition \cite{Frank2021}. It is at least possible that children differ in their cognitive abilities for language and learning, but it is empirically obvious that they differ in their experience with language. Longitudinal records of child language development  have made it possible to track both children's vocabulary growth, and the development of the structural aspects of their grammar. In the Providence Corpus \cite{Demuth2006}, for example, six children were recorded at regular intervals from age 1 to 3. On average, fewer than 20\% of the first 100 words are shared between any two children. The overlap merely rises to about 40\% for the first 1,000, which is the upper limit of a three-year old’s vocabulary size \citep{hart1995meaningful,Bornstein2004}. Yet these children’s grammars are highly uniform even at this stage. Major syntactic categories, word order and argument structure, and the core morphological rules are firmly established before age three \citep{brown1973first} on the basis of at most around 10M words per year \citep{hart1995meaningful} and a vocabulary size of only a few hundred types \citep{Fenson1994}, and all children produce similar grammatical errors during this time.

Recent decades have also seen a convergence between the psychological and formal study of language development and the quantitative study of language variation in early childhood. The sociolinguist, Bill Labov, remarks that ``The end result is a high degree of uniformity in both the categorical and variable aspects of language production, where individual variation is reduced below the level of linguistic significance'' (\citeyear{Labov2012}).

The acquisition of vocabulary and grammar provide clues for investigating the capacities of LMs. Vocabulary learning is a matter of rote learning. This includes not just the arbitrary pairing of sounds and meanings, but also morphological processes (e.g., irregularity) and syntactic structures (e.g., sub-categorization, collocations, etc.). There is no escape from experience: more data results in better learning. But, the structural aspects of the grammar are different. They require form generalizations over the vocabularies.

The distinction between rote learning and structural learning (words vs. rules) is not well reflected by existing LM benchmarks including those discussed in this paper. In practice, these benchmarks are a mixture of tests for both vocabulary learning and grammar learning. Moreover, they are stochastically generated by templates: as such, a large number of test sentences are immediately available, but they lack the structural diversity that has proven revealing in the theoretical study of grammar. 

Furthermore, the sentences are sometimes highly unnatural and semantically/pragmatically uncontrolled, which is precisely the confounding factor that linguists seek to neutralize when attempting to uncover the structural basis of language. \citet{warstadt-etal-2020-blimp-benchmark} are aware that their templates generate unnatural sentences, presenting the BLiMP sentence `\texttt{Sam ran around some glaciers.}' as an example. We found similar issues in Zorro, such as `\texttt{the lie on the foot is flat .},' the first sentence in Zorro's \texttt{across\_prepositional\_phrase} paradigm (\texttt{lie} is a noun). The BLiMP authors state that this is not a problem because it affects both sentences in a pair, but how can we rule out unintended interactions between the grammatical phenomenon under evaluation and the semantic implausibility? \citet{sprouse2018colorless} find that this semantic implausibility may affect judgments of sentence well-formedness, even in the Forced Choice (FC) task used to collect the human baselines in BLiMP.

Indeed, there are already a large amount of carefully curated linguistic materials that are not only structurally diverse but also have minimized lexical and semantic confounds. Furthermore, these datasets (e.g., the Adger/LI dataset; Section \ref{sec:variability}) have been evaluated for acceptability at an individual level by a large pool of native speaker subjects and show very high convergence rates across individuals.  They will be especially informative if we are to explore the structural knowledge of LMs.

\section{Re-examining the Benchmarks}\label{sec:linear}

\subsection*{BLiMP \citep{warstadt-etal-2020-blimp-benchmark}}

\citet{warstadt-etal-2020-blimp-benchmark} introduce the Benchmark of Linguistic Minimal Pairs (BLiMP)\footnote{\url{https://github.com/alexwarstadt/blimp}} as a means of evaluating the linguistic knowledge of neural language models. BLiMP extends the reasoning of earlier studies \citep[e.g.,][]{linzen2016assessing,marvin-linzen-2018-targeted,wilcox-etal-2018-rnn} which use a minimal pair paradigm to approximate acceptability judgments. Instead of prompting for a acceptability judgments on individual sentences, as is most commonly done for human subjects, they present an LM with two sentences that only differ in one structural or lexical property. For a given minimal pair $m_i$ consisting of an acceptable sentence $s_{i,1}$ and an unacceptable sentence $s_{i,2}$, if an LM evaluates $P(s_{i,1}) > P(s_{i,2})$, then the model has succeeded on $m_i$. An LM is scored according to the percentage of all the minimal pairs for which it identified the acceptable sentence.
The minimal pair approach allows for the direct evaluation of LMs without training a binary classifier on top of them as was necessary for previous acceptability benchmarks \citep[e.g., CoLA; ][]{warstadt-etal-2019-neural}.

Minimal pairs need to be carefully constructed to control for length and lexical frequencies. BLiMP aims to accomplish this with automatic generation from templates, but as we discuss, it often yields sentences with low structural diversity and implausible semantics. The benchmark corpus includes data sets for 12 linguistic phenomena, including \textsc{Anaphor agreement}, \textsc{Argument structure}, \textsc{Binding}, \textsc{Control/raising}, and others listed in the Appendix. These are further divided into 67 paradigms, each containing 1000 sentences pairs, which are meant to test variants of the phenomena, for example the phenomenon \textsc{Determiner-noun agr.} contains 6 paradigms for adjacent agreement, agreement with irregular nouns, and agreement with adjectives intervening. BLiMP has become a standard NLP benchmark for this task and will be used as part of the test data for the upcoming BabyLM Challenge.

\subsection*{Zorro \citep{huebner-etal-2021-babyberta}}

\citet{huebner-etal-2021-babyberta} explicitly aim to evaluate the relationship between LMs and the acquisition of grammar. They introduce BabyBERTa\_\aoc\, ``an acquisition-friendly version of RoBERTa,'' trained on English child-directed/produced speech (CDS) approximating the total input of a typical English-learning six-year-old. They train variants on only CDS from \aoc\  \citep{huebner2021using}, a pre-processed version of English CHILDES \cite{macwhinney1991childes}, as well as variants on larger datasets from other sources.

Because BabyBERTa\_\aoc\ (henceforth BabyBERTa) was trained on much less text than typical large transformer models are, its vocabulary is much smaller. To mitigate the impact of out-of-vocabulary (OOV) items on their tests, the authors introduce a new grammaticality test suite, Zorro,\footnote{\url{https://github.com/phueb/Zorro/}} in the style of BLiMP. Sentence pairs are generated for one paradigm each for 11 of BLiMP's 12 phenomena, along with two additional phenomena. However, we show that the Zorro sentences are not only lexically simpler as intended, but their templates are also far less complex and even less varied than the sentences in the corresponding BLiMP phenomena.
Full lists of paradigms for each data set can be found in the Appendix, and the full data sets themselves are made available by the benchmarks' original authors.

\subsection{Linear Baselines}

As noted earlier, BLiMP and Zorro  tests are stochastically generated with category-based templates. This way, a large number of examples can be generated and tested, but the drawback is that all examples are essentially the same structure. Moreover, many of the structures are simple, falling considerably below the coverage of modern syntactic analyses. In fact, many examples appear solvable by strictly linear methods. The observation that such template-generated examples can be solved this way is not new to to field. For example, \citet{kam2008bigrams} demonstrated that a bigram model will predict the grammatical sentence from template-produced pairs featuring auxiliary inversion (a structural phenomenon) as well as neural models of the time. 

To take an example from BLiMP, within its \textsc{Subject-verb agr} phenomenon, four of six paradigms evaluate string-adjacent subject verb agreement that could be captured by a bigram model. The remaining two include intervening distractor nouns, but in both these and the string-adjacent paradigms, the target noun is consistently the first/leftmost noun. A single linear rule, albeit a long-distance one, is sufficient to succeed on this phenomenon. In \textsc{Anaphora agreement}, none of the sentences has any distractors at all: the test is solely about whether the anaphor (e.g., \textit{himself}/\textit{herself}) agrees with the first, and only, noun in the sentence preceding it. Success on such simple tests tells us little about the genuine grammatical capacity of LMs and distorts or dilutes summary metrics calculated over the benchmark.

We evaluate this problem quantitatively with two studies of linear rules that do not incorporate structural knowledge. We find that many, but certainly not all, paradigms are solvable with non-human-like linear approaches. These paradigms therefore do not contribute to the overall goal of evaluating whether an LM possesses linguistic knowledge. Additionally, we find that the paradigms of Zorro tend to be structurally even simpler and less internally varied than the parallel paradigms of BLiMP. It is a weaker benchmark even when accounting for the intended lexical simplicity.

\subsubsection{\textit{N}-Gram Models}

The original BLiMP paper reports the accuracy of a 5-gram model trained on the 3.1B token Gigaword Corpus \citep{graff2003english} in addition to three neural LMs and human performance. They find that the 5-gram model scores above chance (50\%) on all but two phenomena but is outclassed by most of the neural LMs on most paradigms. Performance for all LMs can vary widely across paradigms within one phenomenon. In some cases, there is a clear split between the 5-gram and neural models, suggesting that the latter capture some structural property of the paradigm that the 5-gram model does not, but in other cases, the 5-gram model performs well, demonstrating that linear rules can be sufficient for completing those tasks.

Revisiting \textsc{Subject-verb agr.} as an illustrative example, the Gigaword 5-gram model performs only slightly behind the neural models on each string-adjacent paradigm but far below chance in the distractor paradigms. However, the neural models also perform up to 20.5 points better in the adjacent paradigms than the distractor paradigms. The two distractor paradigms demonstrate that the neural models have learned a long-distance pattern (whether that be structural or ``agree with the leftmost noun''), but the adjacent paradigms cannot show this. They, and about half of the BLiMP paradigms, are uninformative in this way.

We extend this approach to the language acquisition setting by training a 5-gram model only on \aoc\ and evaluating on both BLiMP and Zorro. We compare these results to BabyBERTa on these data sets.\footnote{Refer to Appendix for full details. We downloaded the publicly available \href{https://github.com/phueb/BabyBERTa/tree/master/saved_models}{model checkpoints} from the BabyBERTa GitHub repository and replicated the \href{https://github.com/phueb/UnMasked/tree/master/results}{BLiMP and Zorro results} hosted on the Zorro GitHub repository} To further manage lexical effects while adding minimal complexity to the model, we evaluate both a 5-gram word model (5-word), and a 5-gram model trained only on POS tags (5-tag). \aoc\ was tagged using GPoSTTL, a rule-based POS tagger with tokenizer and lemmatizer based on the Brill Tagger \cite{brill1992}. This was used to train sklearn's CRF POS-tagger, which was then used to label the benchmark corpora. This approach was taken to avoid bringing additional knowledge from a tagger trained on larger corpora into the benchmark corpora. The downside is that the tagger is not particularly accurate on the ungrammatical benchmark sentences, which may hurt performance for the 5-tag model. In addition to the 5-word and 5-tag models, we evaluate an oracle which marks a correct prediction if either 5-word or 5-tag makes a correct prediction.  Our  use of POS is motivated from a developmental perspective. Syntactic categories can be formed purely distributionally as early as infancy \cite{Mintz2003,Shi2010, Reeder2013} and children almost never make mistakes in their use of syntactic categories \cite{Valian1986}. It is thus plausible to assume that the acquisition of grammatical knowledge builds on a developmentally prior stage of syntactic category learning. 

The results of the 5-gram experiments are summarized in Table \ref{tab:ngramsummary} and laid out in detail in the Appendix. We draw three conclusions from these. First, the 5-gram models perform surprisingly well relative to the BabyBERTa transformer despite its extremely non-human-like simplicity when trained on the same \aoc\ data. Either 5-word or 5-tag, trained on the same data as BabyBERTa, outperformed BabyBERTa on 11 of 23 Zorro paradigms and 21 of 67 BLiMP paradigms. BabyBERTa's performance appears less impressive when presented alongside even this very weak baseline. The \aoc\ 5-gram models perform more poorly on BLiMP than the Gigaword 5-gram model, but it still achieves high accuracy on several paradigms scattered across the phenomena.

Second, 5-gram oracle outperforms 5-word, 5-tag, and BabyBERTa. The 5-gram oracle is not a fair direct comparison but provides a summary metric for correlation between 5-word and 5-tag. A high oracle score relative to the two 5-gram models indicates that they do not make the same errors. That is, errors are not necessarily attributable to the string-local limitations of 5-grams \textit{per se} but rather to 5-gram sparsity or errors in tagging. The high oracle score is another sign that the paradigms often capture surface properties rather than structural properties that would stump 5-gram models.

Third, the 5-gram models outperform BabyBERTa on proportionately more Zorro paradigms than BLiMP paradigms. Additionally, the \aoc\ 5-word model achieved 78.91\% performance on Zorro, while the Gigaword 5-gram model only reached 60.5\% on BLiMP. If Zorro were merely accounting for the smaller vocabulary in the \aoc\ training data, we should expect much more similar performance on both of these measures. Taken together, these suggest that Zorro is a substantially weaker benchmark that BLiMP, and it more greatly overestimates the apparent positive results of the acquisition-inspired BabyBERTa.

\begin{table}[]
    \centering
    \scriptsize
    \begin{tabular}{c|c|cc|cc}
    \hline
    \textbf{Zorro} & \textbf{BabyBERTa} & \textbf{5-Word} & \textbf{5-Tag} & \textbf{Either} & \textbf{Oracle}\\
    \hline
    \textbf{\# Best} & \--- & 8/23 & 8/23 & 11/23 & 14/23\\
    \textbf{Avg Acc} & 78.91\% & 63.44\% & 57.59\% & \--- & 83.43\%\\
    \hline
    \hline
    \textbf{BLiMP} & \textbf{BabyBERTa} & \textbf{5-Word} & \textbf{5-Tag} & \textbf{Either} & \textbf{Oracle}\\
    \hline
    \textbf{\# Best} & \--- & 18/67 & 10/67 & 23/67 & 48/67\\
    \textbf{Avg Acc} & 60.72\% & 50.72\% & 37.93\% & \--- & 68.32\% \\
    \end{tabular}
    \caption{Performance summaries for 5-grams relative to BabyBERTa on Zorro and BLiMP. Number of paradigms in which a 5-gram model outperforms BabyBERTa and overall average accuracy across paradigms are reported. Either = either 5-word or 5-tag outpeformed BabyBERTa on the entire paradigm. Oracle = sentence pairs were marked correct if either 5-word or 5-tag made the correct prediction.}
    \label{tab:ngramsummary}
\end{table}

\subsubsection{Hand-Written Simple Rules}

In addition to reporting results on 5-gram models, we created simple hand-written rules which demonstrate that the probes are solvable in principle without reference to linguistic structure. While we do not claim that such rules are akin to the state of knowledge in LMs, it is also difficult to completely rule out this possibility. On the one hand, it is still unclear how to interpret the representation of linguistic knowledge in LMs. On the other, the vast majority of training data, at least child-directed for language acquisition, is structurally simple and can in fact be handled by rule-like pattern matchers. In English CDS, the distribution of anaphora is exceedingly straightforward: almost all instances of {\it himself} are preceded in the sentence by the subject pronoun {\it he} and a (male) noun phrase with no co-referential  competitors. For comparison, Zorro \texttt{adjunct\_island} can be solved perfectly by always selecting the sentence where the third-last word is \textit{the}, and many of the paradigms can be solved by tracking the index of a specific word. Others can be solved by checking for the presence of a certain word. For example, the \texttt{superlative} paradigm can be solved by accepting the sentence that contains either \textit{more} or \textit{fewer}. For both Zorro and BLiMP, more than one paradigm can often be solved with the exact same rule. We write  simple linear rules for each Zorro and BLiMP paradigm. See the Appendix for a full list of rules. 

In summary, these rules yielded 93.97\% accuracy on Zorro and solved 14 of 23 Zorro paradigms with 100\% accuracy. Each \texttt{agreement\_} paradigm is solved with at least 96\% accuracy, with the remainder due to two irregular nouns, \textit{feet} and \textit{children}, which do not end in the \textit{-s} referenced by these rules. The lowest performance is 52.75\% on \texttt{anaphor\_agreement-pronoun\_gender}, a paradigm that requires an LM to `know' the canonical gender of English names in order to choose \textit{himself} or \textit{herself}. The test sentence pairs were not quite balanced, so always guessing \textit{himself} earns more than 50\%.

BLiMP proved more challenging. The rules only yielded 84.35\% accuracy on average and achieved perfect scores on 14 of 67 rules. The overall high score of the hand-written simple linear rules suggests that BLiMP suffers from the same issues regarding lack of sentence variety that Zorro does, but the lower accuracy indicates that the problem is not quite as severe. In principle, we could have composed more complex rules which achieved perfect accuracy on all paradigms, however, these simpler rules better illustrate our points. The success of non-human-like simple linear rules on most paradigms on both benchmarks further emphasizes that success on the template-based behavioral task does not necessarily imply that an LM possesses linguistic knowledge.

\begin{table*}[h!]
    \centering
    \small
    \begin{tabular}{llc}
    \hline
    \textbf{Sentence ID} & \textbf{Sentence} & \textbf{ME Z-score} \\
    \hline
    32.3.Culicover.7a.g.01 & John tried to win. & 1.453262 \\
    32.3.Culicover.7b.*.01 & John tried himself to win. & -0.86729 \\
    33.2.bowers.7b.g.07 & Sarah counted the change accurately. & 1.230412 \\
    33.2.bowers.7b.*.07 & Sarah accurately counted the change. & 1.20698 \\
    \hline
    ch8.150.*.01 & Melissa seems that is happy. & -1.14131 \\
    ch8.151.g.01 & It seems that Melissa is happy. & 1.000644 \\
    ch8.152.g.01 & Melissa seems to be happy. & 1.196088\\
    \hline
    \end{tabular}
    \caption[Example phenomena from LI-Adger dataset]{Top: Two pairwise phenomena from the Linguistic Inquiry (LI) dataset.  Bottom: One multi-condition phenomenon from the Adger dataset.  The ME Z-score is the averaged Z-score transformation of the human Magnitude Estimation judgments for each of the  sentences across all the experimental participants.}
    \label{tab:human_me_example}
\end{table*}

\section{An Alternative: The LI-Adger Dataset}\label{sec:variability}

The LI-Adger dataset is a comprehensive collection of 519 sentence types, 300 collected by \citet{sprouse2013comparison} from \textit{Linguistic Inquiry (LI) 2001-2010},\footnote{\url{https://www.jonsprouse.com/data/Lingua2013/SSA.data.zip}} a major theoretical journal in linguistics, and 219 collected by \citet{sprouse2012assessing} from \citeposs{adger2003core} \textit{Core Syntax} textbook.\footnote{\url{https://www.jonsprouse.com/data/JoL2012/SA2012.data.zip}}  Each sentence type includes eight hand-constructed, semantically plausible sentences, assembled into 150 pairwise (LI) and 105 multi-condition (Adger) phenomena where each minimal pair is lexically matched.  We provide an example of each in Table \ref{tab:human_me_example}.

The LI-Adger dataset improves upon the prior two datasets in three key ways.  Firstly, unlike BLiMP and Zorro, the LI-Adger sentences are controlled for semantic implausibility, which has been shown to be a strong confounding factor when eliciting human judgments \citep{sprouse2018colorless}. Second, the 255 total pairwise and multi-condition phenomena achieve much more diverse coverage of syntactic phenomena than the 67 paradigms in BLiMP, and the 23 paradigms in Zorro.  Third, the human judgments were collected using the Magnitude Estimation (ME) task (and Likert Scale (LS) in the case of the LI sentences) in addition to Forced-Choice (FC) as in the BLiMP human baselines.  We believe this to be a crucial advantage because the FC task treats sentence acceptability as functionally categorical: A sentence is only acceptable or not relative to its minimal pair counterpart, whereas tasks such as ME allow us to make comparisons within and across minimal pairs, thereby treating sentence acceptability as a gradient measure. 

With this dataset, we conduct the following two tests.  First, in line with \citet{vazquez-martinez-2021-acceptability}, we sort the LI-Adger dataset into 2391 unique minimal pairs.  We then collect pseudo log-likelihood scores for each sentence from several models evaluated by \citet{huebner-etal-2021-babyberta}, and score them using the same criteria as BLiMP and Zorro.  As a baseline for the models, we include Log-Likelihood and Syntactic Log-Odds Ratio (SLOR; \citealp{pauls2012large, lau2017grammaticality}) scores by a trigram model trained on the British National Corpus (BNC; 100M words) by \citet{sprouse2018colorless}.

We include the results of this test in Figure \ref{fig:li-adger_performance}.  We observe that all models are further from the human baseline as compared to those in BLiMP (no human baselines were reported for Zorro).  But more importantly, we observe that the trigram model scored using SLOR performs on par with the BabyBERTa models and approaches the performance of RoBERTa \citep{liu2019roberta} trained on 10M words. If we were to adopt the ``positive results from model learners are more meaningful than negative results'' argument, then the trigram model is as suitable a model of language acquisition as BabyBERTa is.

\begin{figure*}[t]
    \centering
    \includegraphics[width=7.5cm]{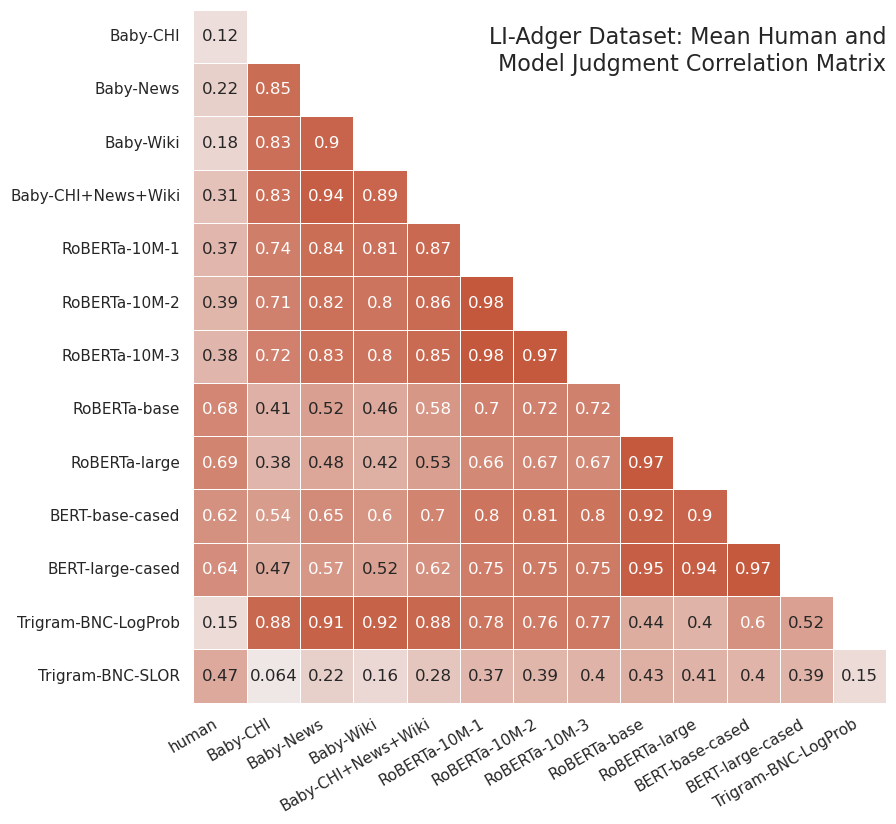}
    \includegraphics[width=7.5cm]{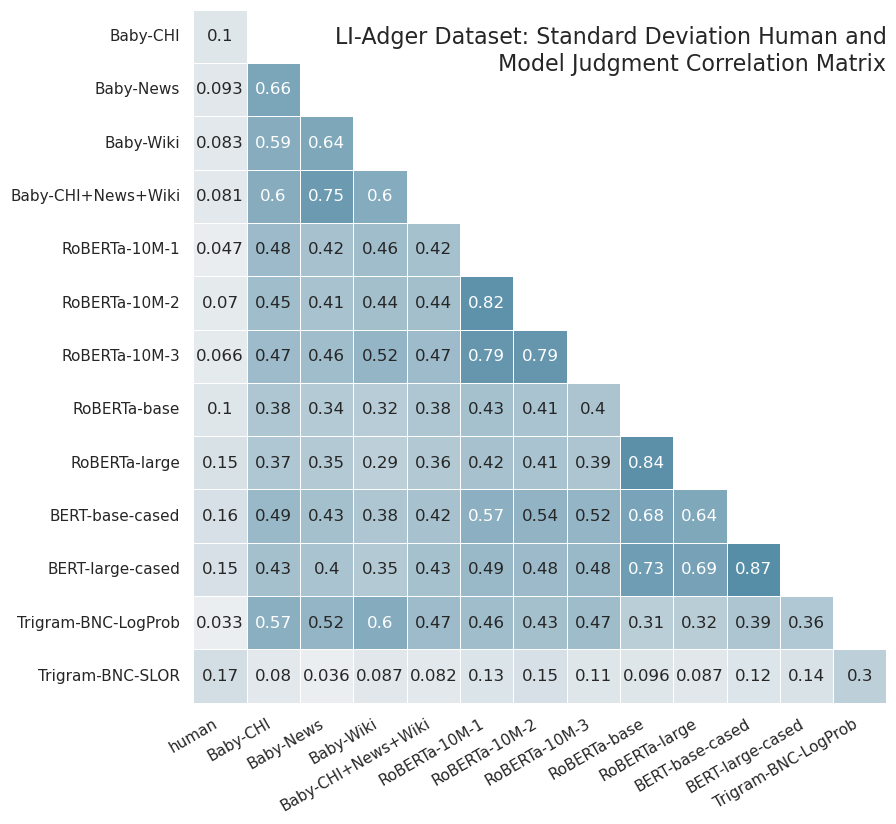}
    \vspace{-0.3cm}
    \caption{Correlation matrices of human judgments and LM output means (top) and standard deviations (bottom) on each sentence type on the LI-Adger dataset. Baby=BabyBERTa, CHI=\aoc, News=\textsc{ao-newsela}, Wiki=Wikipedia-1.}
    \label{fig:li-adger-corr}
\end{figure*}

Raw accuracy notwithstanding, we proceed to conduct a novel test of judgment variability on our collection of LMs.  We take advantage of the structure of the LI-Adger dataset in the following way: There are 519 sentence types, and for each type there are eight sentences that retain the same syntactic structure but vary lexical items at the locus of the syntactic structure tested (e.g., the head of a verb phrase from which extraction takes place). These datasets thus allow us to contrast the consistency of human judgment across and within construction types against that of the LMs.

We z-score  the LM judgments to make them comparable to the human judgments. Then, for each set of eight sentences, we take the mean and standard deviation of all the judgments for humans and each LM.  We find that the models are much more variable in their judgments: The human judgments, on average, vary by 0.288 standard deviation (std. dev.) units within a given set of sentences.  On the other hand, the LM that least varies is BabyBERTa Wiki, varying by 0.451 std. dev. units on average.  The rest of the models nearly double the variability of the human judgments, ranging from 0.518 for RoBERTa-10M-1 to 0.554 for BERT-large-cased. Variability appears to increase rather than decrease as training size and performance increase. Surprisingly, the trigram model, when scored using log probabilities, is the closest in variability to the human judgments at 0.331 std. dev. units, but also the furthest when scored using SLOR with a variability of 0.599.  Once again we find that a positive result on one test or another is not enough to draw positive conclusions.

For further illustration, we correlate the means and standard deviations of 512 sentence types across each LM and humans and plot the results in Figure \ref{fig:li-adger-corr}. Both in terms of mean and standard deviations, we observe generally high correlations between the various neural LMs, but much lower correlations between the LMs and humans. This suggests that whatever the LMs are doing, good or bad, does not appear to be human. Interestingly, the BabyBERTa LMs show very high correlations with the naive trigram log-likelihood scores and very low with trigram SLOR scores, raising further suspicions that these small acquisition-inspired LMs behave like a very non-human-like model.

\section{Discussion}\label{sec:discussion}

It is widely recognized that children acquire language in ways that appear quite different from LM training. There is a growing realization that the cognitive relevance of LMs can only be established in a comparable setting. Bringing down training size requirements stands not to not only improve the applicability of such models to the study of language acquisition but also to efficient NLP on low-resource languages. 

However, in this paper, we observed several weaknesses in BLiMP and Zorro, two minimal pair benchmarks for evaluating the linguistic knowledge of neural language models. We believe that it is worth critically revisiting the underlying assumption that positive results on such benchmarks are a demonstration of human-like representations or human-like language acquisition, especially if  an evaluation can be solved in unintended ways, or if it does not reflect an adequately broad range of linguistic structures. 
It is unlikely that a behavioral probe, such as these large binarized  benchmarks, can  fully capture the complexity of linguistic knowledge. To this end, we made a case for also evaluating with curated benchmarking datasets: the gradient acceptability judgments from human subjects makes these  effective probes for the structural basis of grammar. Together with a range of tests, from carefully constructed tests of grammaticality to probes correlating the internal state of LMs with their predictions need to complement theoretical, psycholinguistic, and neurolinguistic studies before a meaningful cognitive claim about the nature of neural language models can be made.

We end with some broader discussion about language acquisition and the cognitive interpretation of computational models. While it is now widely recognized that children learn language with only a fraction of the data needed for large LM training, merely reducing the amount of training data alone\,--\,such as the 100M word threshold in the BabyLM Challenge\,--\,still falls short of the requirement for an adequate model of language acquisition. 
While it is true that a native speaker's knowledge of language can be established on the basis of approximately 100 million words, child language research makes clear that not all aspects of linguistic knowledge are learned at the same time. Some, such as inflectional morphology, case marking, word order, and major transformations are acquired very early in all languages studied so far \citep[e.g.,][]{brown1973first, Slobin2022} at an order of magnitude fewer words of input, while others are learned rather late: These include derivational morphology \cite{Jarmulowicz2002}, passivization \cite{Pinker1987}, control and cleft structures \cite{Chomsky1969} and the dative constructions \cite{Gropen1989} in the case of English, but these may emerge much earlier in other languages. This suggests that successful learning in the limit (e.g., 100M word) is not sufficient. For example, while a neural model of English past tense \cite{Kirov2018} eventually learns the "add -ed" rule, it does so with over 3,000 verb lemmas. By contrast, children learn that rule before or around 3 \cite{Kuczaj1977}, when their vocabulary only contains around 300 or so verbs \citep{marcus1992overregularization}. To serve as cognitive models of language, it is thus important to compare the training trajectory of LMs as a function of the training data volume against the developmental benchmarks of specific linguistic phenomena which have been amply documented in the empirical literature on child language.

\section*{Limitations}

Our study is about the limitations of evaluation, so it is to be expected that our study has its limits as well. Most obviously, ours and any study would benefit from testing and reporting on a wider range of neural models and a wider range of baselines. And like most work in this area, our evaluations were only performed on English. We recommend the use of the LI-Adger data set. Like any behavioral probe, including the ones which we criticize, it can be subject to ambiguous interpretation. It has some substantial advantages, as we discuss in this paper, but also a couple of additional drawbacks. It is smaller than BLiMP or Zorro, and it has not been annotated by phenomenon. Nevertheless, it provides additional insights that those benchmarks do not. As in the paper, we recommend its use in conjunction with a wide range of other evaluation methods.


\bibliography{anthology,main}
\bibliographystyle{acl_natbib}

\onecolumn
\section*{Appendix}
\label{sec:appendix}

\begin{table*}[h!]
    \centering
    \small
    \begin{tabular}{ll|c|ccc|c}
    \hline
Phenomenon & Paradigm & BabyBERTa &\multicolumn{3}{|c|}{5-Gram} & Simple\\
& & \textsc{ao-childes} & Word & Tag & Oracle & Rule\\
    \hline
\texttt{agreement\_subject\_verb} & \texttt{across\_rel\_clause} & 64.85 & 50.95 & 46.35 & \textbf{68.95} & \textbf{96.20}\\
& \texttt{in\_simple\_question} & 92.35 & 61.15 & 90.9 & \textbf{93.9} & \textbf{98.30}\\
& \texttt{in\_question\_with\_aux} & 90.85 & 59 & 80.15 & \textbf{90.9} & \textbf{98.05}\\
 & \texttt{across\_prep\_phrase} & 72.85 & 50 & 50 & 62.6 & \textbf{98.40}\\
\texttt{agreement\_determiner\_noun} & \texttt{between\_neighbors} & 91.3 & 83.05 & 49.85 & 88.6 & \textbf{98.60}\\
 & \texttt{across\_1\_adjective} & 89.85 & 50.45 & 50.05 & 75.05 & \textbf{97.20}\\
\texttt{filler-gap} & \texttt{wh\_question\_object} & 98.75 & 42.8 & \textbf{100} & \textbf{100} & \textbf{100}\\
& \texttt{wh\_question\_subject} & 75.7 & \textbf{88.3} & \textbf{76.55} & \textbf{97.1} & \textbf{100}\\
\texttt{island-effects} & \texttt{coord\_struct\_constr} & 97.05 & 43.35 & 55.6 & 83.85 & \textbf{100}\\
& \texttt{adjunct\_island} & 56.15 & \textbf{66.1} & \textbf{58.8} & \textbf{83.85} & \textbf{100}\\
\texttt{quantifiers} & \texttt{existential\_there} & 92.9 & 80.25 & 38.4 & 89.55 & \textbf{100}\\
& \texttt{superlative} & 64.55 & 45.1 & \textbf{82} & \textbf{96.05} & \textbf{100}\\
\texttt{npi\_licensing} & \texttt{only\_npi\_licensor} & 74.1 & \textbf{79.4} & 3.7 & \textbf{79.4} & \textbf{100}\\
& \texttt{matrix\_question} & 65.25 & 47.5 & 28.65 & 58 & \textbf{100}\\
\texttt{argument\_structure} & \texttt{swapped\_arguments} & 91 & \textbf{92.15} & 81.7 & \textbf{98.85} & \textbf{100}\\
 & \texttt{transitive} & 60.05 & \textbf{64.15} & 32.65 & \textbf{78.6} & 58.05\\
& \texttt{dropped\_argument} & 79.9 & \textbf{85.05} & \textbf{83.6} & \textbf{95.75} & \textbf{100}\\
\texttt{irregular} & \texttt{verb} & 69.65 & 62.9 & \textbf{93.6} & \textbf{96.35} & \textbf{88.40}\\
\texttt{anaphor\_agreement} & \texttt{pronoun\_gender} & 51.75 & 49.15 & 1.95 & 50.95 & \textbf{52.75}\\
\texttt{ellipsis} & \texttt{n\_bar} & 55.3 & \textbf{66.6} & \textbf{63.6} & \textbf{89.9} & \textbf{100}\\
\texttt{binding} & \texttt{principle\_a} & 89.4 & 45.9 & 3.6 & 47.75 & \textbf{100}\\
\texttt{case} & \texttt{subjective\_pronoun} & 94.7 & \textbf{99.55} & \textbf{97.95} & \textbf{100} & \textbf{100}\\
\texttt{local\_attractor} & \texttt{in\_question\_with\_aux} & 96.65 & 55.65 & 95 & \textbf{99.05} & \textbf{100}\\
\hline
\multicolumn{2}{l|}{\textsc{average}} & 78.91\% & 63.44\% & 57.59\% & 83.43\% & 93.97\%\\
\multicolumn{2}{l|}{Fraction $\geq$ BabyBERTa} & \--- & 8/23 & 8/23 & 14/23 & 22/23\\
    \end{tabular}
    \caption{Word and tag-level 5-gram models trained on \textsc{ao-childes} plus 5-Gram Oracle and Simple Linear Rule Oracle for Zorro. 5-Gram and Simple Rule scores that are greater than BabyBERTa\_\aoc\ are bolded}
    \label{tab:zorro-superficial}
\end{table*}

\begin{table*}[]
    \centering
    \small
    \begin{tabular}{ll|l}
    \hline
Phenomenon & Paradigm & Rule\\
    \hline
\texttt{agreement\_subject\_verb} & \texttt{across\_rel\_clause} & 2nd word ends in \textit{s} iff 3rd last is in \{\textit{are}, \textit{were}, \textit{do}\}\\
& \texttt{in\_simple\_question} & Word 2 right of \{\textit{are}, \textit{were}\} ends in \textit{s}.\\
& & Word 2 right of \{\textit{is}, \textit{are}\} does not\\
& \texttt{in\_question\_with\_aux} & 4th word ends in \textit{s} iff 2nd is in \{\textit{are}, \textit{were}, \textit{do}\}\\
 & \texttt{across\_prep\_phrase} & 2nd word ends in \textit{s} iff 3rd last is in \{\textit{are}, \textit{were}, \textit{do}\}\\
\texttt{agreement\_determiner\_noun} & \texttt{between\_neighbors} & If \{\textit{these}, \textit{those}\} in sentence, next word ends in \textit{s}. \\
& & If \{\textit{this}, \textit{that}\} in sentence, next word does not\\
 & \texttt{across\_1\_adjective} & If \{\textit{these}, \textit{those}\} in sentence, word 2 right ends in \textit{s}.\\
 & & If \{\textit{this}, \textit{that}\} in sentence, word 2 right does not\\
\texttt{filler-gap} & \texttt{wh\_question\_object} & 2nd word is \textit{the}\\
& \texttt{wh\_question\_subject} & \textit{who} does not immediately precede \textit{the}\\
\texttt{island-effects} & \texttt{coord\_struct\_constr} & 4th word is \textit{and}\\
& \texttt{adjunct\_island} & 3rd last word is \textit{the}\\
\texttt{quantifiers} & \texttt{existential\_there} & Contains one of \{\textit{many}, \textit{some}, \textit{no}, \textit{few}, \textit{a}, \textit{an}\}\\
& \texttt{superlative} & Contains one of \{\textit{more}, \textit{fewer}\}\\
\texttt{npi\_licensing} & \texttt{only\_npi\_licensor} & 1st word is \textit{only}\\
& \texttt{matrix\_question} & Contains one of \{\textit{does}, \textit{will}, \textit{should}, \textit{could}, \textit{did}, \textit{wouldo}\}\\
\texttt{argument\_structure} & \texttt{swapped\_arguments} & 1st word is \textit{the}\\
 & \texttt{transitive} & 2nd last word does not end in \textit{e}\\
& \texttt{dropped\_argument} & 1st word is \textit{the}\\
\texttt{irregular} & \texttt{verb} & word following \textit{had} ends in \textit{n} or no word ends in \textit{n}\\
\texttt{anaphor\_agreement} & \texttt{pronoun\_gender} & Sentence contains \textit{himself}\\
\texttt{ellipsis} & \texttt{n\_bar} & *Sentence where \textit{and} appears farther right\\
\texttt{binding} & \texttt{principle\_a} & 4th last word ends with \textit{ing}\\
\texttt{case} & \texttt{subjective\_pronoun} & 1st word is \textit{the} \\
\texttt{local\_attractor} & \texttt{in\_question\_with\_aux} & 4th word does not end with \textit{'s}\\

    \end{tabular}
    \caption{Simple Linear Rule descriptions for Zorro. Rules that require sentences to be compared are marked with an asterisk.}
    \label{tab:zorro-rules}
\end{table*}

\begin{table*}[]
    \centering
    \scriptsize
    \begin{tabular}{ll|c|ccc|c}
    \hline
Phenomenon & Paradigm & BabyBERTa &\multicolumn{3}{|c|}{5-Gram} & Simple\\
& & \textsc{ao-childes} & Word & Tag & Oracle & Rule\\
    \hline
\texttt{anaphor\_agreement} & \texttt{anaphor\_gender\_agreement} & 65.6 & 26.3 & 8 & 33.9 & \textbf{73.9}\\
& \texttt{anaphor\_number\_agreement} & 73.7 & 52.9 & 5.7 & 55.5 & \textbf{80.1}\\
\texttt{argument\_structure} & \texttt{causative} & 58.5 & 55.2 & 30.7 & \textbf{68.8} & \textbf{85.6}\\
 & \texttt{drop\_argument} & 63.2 & 50.9 & 52.9 & \textbf{80.8} & \textbf{77.1}\\
& \texttt{inchoative} & 50.7 & \textbf{56} & 37.1 & \textbf{73.8} & \textbf{57.1}\\
& \texttt{intransitive} & 52.1 & 48.2 & 49.6 & \textbf{76.3} & \textbf{73.55}\\
 & \texttt{passive\_1} & 50.2 & \textbf{52.1} & 12.9 & \textbf{56.4} & \textbf{59.5}\\
& \texttt{passive\_2} & 54 & \textbf{48.4} & 18.1 & \textbf{56.8} & \textbf{59.6}\\
& \texttt{transitive} & 55.3 & 51.6 & 36.1 & \textbf{67.6} & \textbf{57.85}\\
\texttt{binding} & \texttt{principle\_A\_case\_1} & 43.6 & \textbf{100} & 7.1 & \textbf{100} & \textbf{100}\\
& \texttt{principle\_A\_case\_2} & 99.9 & 41.5 & 13 & 48.3 & 99.2\\
 & \texttt{principle\_A\_c\_command} & 58.7 & 35.7 & 4.2 & 38.1 & \textbf{71.35}\\
& \texttt{principle\_A\_domain\_1} & 96.5 & 38.4 & 3.1 & 40.7 & \textbf{100}\\
 & \texttt{principle\_A\_domain\_2} & 51.4 & \textbf{61.7} & 2.7 & \textbf{62.8} & \textbf{58.3}\\
& \texttt{principle\_A\_domain\_3} & 46.8 & 44.5 & 29.7 & \textbf{61.1} & \textbf{50.4}\\
 & \texttt{principle\_A\_reconstruction} & 40.9 & 32.1 & \textbf{53.9} & \textbf{68} & \textbf{74.1}\\
\texttt{control\_raising} & \texttt{existential\_there\_object\_raising} & 59.1 & 30.5 & 23.4 & 46.5 & \textbf{67.95}\\
& \texttt{existential\_there\_subject\_raising} & 51 & 43.4 & 17 & \textbf{53.6} & \textbf{77}\\
 & \texttt{expletive\_it\_object\_raising} & 63.3 & 61.2 & 48.3 & \textbf{79.6} & \textbf{69.5}\\
& \texttt{tough\_vs\_raising\_1} & 72.2 & 59.1 & 49.6 & \textbf{83.2} & \textbf{87.1}\\
& \texttt{tough\_vs\_raising\_2} & 34.4 & \textbf{41.3} & 18.4 & \textbf{54.1} & \textbf{92.5}\\
\texttt{determiner\_noun\_} 
& \texttt{a\_irregular\_1} & 66.6 & 48.8 & 37.4 & 61.3 & \textbf{68.45}\\
\texttt{agreement = a} & \texttt{a\_irregular\_2} & 87.4 & 74.3 & 12.3 & 77.1 & 73.7\\
& \texttt{a\_with\_adjective\_1} & 76.3 & 48.2 & 49.7 & 63.8 & \textbf{95.95}\\
& \texttt{a\_with\_adj\_irregular\_1} & 82.9 & 49 & 49.7 & 56.3 & 74.45\\
& \texttt{a\_with\_adj\_irregular\_2} & 67 & 49.5 & 18.3 & 58.2 & \textbf{71.8}\\
& \texttt{a\_with\_adj\_2} & 80.4 & 49.8 & 19.9 & 59.7 & \textbf{95.6}\\
& \texttt{a\_1} & 72.2 & 64.1 & 48.1 & \textbf{74.5} & \textbf{95.55}\\
& \texttt{a\_2} & 87.4 & 65.2 & 11 & 68.1 & \textbf{96.75}\\
\texttt{ellipsis} & \texttt{ellipsis\_n\_bar\_1} & 58.7 & \textbf{64.1} & \textbf{63.5} & \textbf{86.4} & \textbf{85.65}\\
& \texttt{ellipsis\_n\_bar\_2} & 42.8 & 39.9 & \textbf{70.5} & \textbf{80.9} & \textbf{99.95}\\
\texttt{filler\_gap\_dependency} & \texttt{wh\_questions\_object\_gap} & 73 & 37 & \textbf{82.4} & \textbf{89.2} & \textbf{99.95}\\
& \texttt{wh\_questions\_subject\_gap} & 79.9 & 49 & \textbf{81.4} & \textbf{89.4} & \textbf{99.9}\\
& \texttt{wh\_vs\_that\_no\_gap} & 90.9 & 77.2 & 83.8 & \textbf{94.9} & \textbf{99.95}\\
& \texttt{wh\_vs\_that\_no\_gap\_long\_distance} & 92.1 & 74.9 & 87 & \textbf{95.8} & \textbf{99.7}\\
& \texttt{wh\_vs\_that\_with\_gap} & 29.1 & 22.7 & 15 & \textbf{33} & \textbf{100}\\
& \texttt{wh\_vs\_that\_with\_gap\_long\_distance} & 14.9 & \textbf{25.8} & 12.8 & \textbf{32.8} & \textbf{99.9}\\
\texttt{irregular\_forms} & \texttt{irregular\_past\_participle\_adjectives} & 59.8 & \textbf{99.4} & 12.2 & \textbf{99.4} & \textbf{100}\\
& \texttt{irregular\_past\_participle\_verbs} & 59.8 & \textbf{99.4} & 12.2 & \textbf{99.4} & \textbf{100}\\
\texttt{island\_effects} & \texttt{adjunct\_island} & 63.8 & 58.4 & 55.5 & \textbf{82.5} & \textbf{94.5}\\
\texttt{(coordinate\_structure\_} & \texttt{y\_complex\_left\_branch} & 36.2 & 11.8 & 19.6 & 26.9 & \textbf{97.05}\\
\texttt{constraint = y)} & \texttt{y\_object\_extraction} & 56.5 & 41.9 & 37.1 & \textbf{63.7} & \textbf{86.35}\\
& \texttt{left\_branch\_island\_echo\_question} & 52.4 & 16.3 & 30.1 & 38.7 & \textbf{100}\\
& \texttt{left\_branch\_island\_simple\_question} & 66.6 & 24.5 & 30.3 & 43.8 & \textbf{97.9}\\
& \texttt{sentential\_subject\_island} & 46.1 & 37.3 & 42.8 & \textbf{62.9} & \textbf{82.65}\\
& \texttt{wh\_island} & 47.1 & \textbf{69} & \textbf{93.4} & \textbf{97.3} & \textbf{100}\\
\texttt{npi\_licensing} & \texttt{matrix\_question\_npi\_licensor\_present} & 56.4 & 41.1 & 39.5 & \textbf{65.7} & \textbf{97.4}\\
& \texttt{npi\_present\_1} & 27 & \textbf{56} & 26.7 & \textbf{69.6} & \textbf{100}\\
& \texttt{npi\_present\_2} & 20.3 & \textbf{56.4} & \textbf{25.8} & \textbf{70.5} & \textbf{100}\\
& \texttt{only\_npi\_licensor\_present} & 71.6 & \textbf{98.4} & 2.4 & \textbf{98.5} & \textbf{100}\\
& \texttt{only\_npi\_scope} & 72.1 & \textbf{80.4} & \textbf{79.4} & \textbf{97.2} & \textbf{100}\\
& \texttt{sentential\_negation\_npi\_licensor\_present} & 73.8 & \textbf{100} & 0 & \textbf{100} & \textbf{100}\\
& \texttt{sentential\_negation\_npi\_scope} & 81.9 & 40 & 65.3 & 79.6& \textbf{100}\\
\texttt{quantifiers} & \texttt{existential\_there\_quantifiers\_1} & 93.7 & 79.1 & 26.4 & 87.4 & \textbf{97.3}\\
& \texttt{existential\_there\_quantifiers\_2} & 35.7 & 19.6 & \textbf{36} & \textbf{50.6} & \textbf{96.85}\\
& \texttt{superlative\_quantifiers\_1} & 49.5 & \textbf{73} & \textbf{89.8} & \textbf{96.4} & \textbf{100}\\
& \texttt{superlative\_quantifiers\_2} & 61.2 & 51.9 & 0.1 & 52 & \textbf{100}\\
\texttt{s-selection} & \texttt{animate\_subject\_passive} & 45.5 & \textbf{48.4} & 24 & \textbf{58.4} & \textbf{65.25}\\
& \texttt{animate\_subject\_trans} & 59.7 & 50 & 57.1 & \textbf{78.2} & \textbf{84.65}\\
\texttt{subject\_verb\_agreement} & \texttt{distractor\_agreement\_relational\_noun} & 29 & 26.2 & 21.4 & \textbf{42.1} & \textbf{50.25}\\
& \texttt{distractor\_agreement\_relative\_clause} & 35.6 & 28.3 & 30.4 & \textbf{49.8} & \textbf{55.85}\\
& \texttt{irregular\_plural\_subject\_verb\_agreement\_1} & 67.9 & 33.4 & 51.7 & 62.5 & 53.2\\
& \texttt{irregular\_plural\_subject\_verb\_agreement\_2} & 66.2 & 51 & 51.9 & \textbf{70.7} & 59.3\\
& \texttt{regular\_plural\_subject\_verb\_agreement\_1} & 68.8 & 39.9 & 51.1 & \textbf{72} & 64.35\\
& \texttt{regular\_plural\_subject\_verb\_agreement\_2} & 60.1 & 51 & 55.6 & \textbf{76.9} & \textbf{73.15}\\
\hline
\multicolumn{2}{l|}{\textsc{average}} & 60.72\% & 50.72\% & 37.93\% & 68.32\% & 84.35\%\\
\multicolumn{2}{l|}{Fraction $\geq$ BabyBERTa} & \--- & 18/67 & 10/67 & 48/67 & 62/67\\
    \end{tabular}
    \caption{Word and tag-level 5-gram models trained on \textsc{ao-childes} plus 5-Gram Oracle and Simple Linear Rule Oracle for BLiMP. 5-Gram and Simple Rule scores that are greater than BabyBERTa\_\aoc\ are bolded}
    \label{tab:blimp-superficial}
\end{table*}

\begin{table*}[]
    \centering
    \scriptsize
    \begin{tabular}{ll|l}
    \hline
Phenomenon & Paradigm & Rule\\
    \hline
\texttt{anaphor\_agreement} & \texttt{anaphor\_gender\_agreement} & Does not contain \textit{itself}\\
& \texttt{anaphor\_number\_agreement} & Number of words that end in \textit{s} is even\\
\texttt{argument\_structure} & \texttt{causative} & Does not contain one of \{\textit{appear}, \textit{vanish}, \textit{exist},\\
& \texttt{transitive} & \textit{sigh}, \textit{rust}, \textit{cheer}, \textit{clash}, \textit{fall}, \textit{fell}, \textit{waste}\}\\
 & \texttt{drop\_argument} & Last word is not one of \{\textit{to},\\
& \texttt{inchoative} & \textit{with}, \textit{about}, \textit{from}, \\
& \texttt{intransitive} & \textit{at}, \textit{through}, \textit{by}, \textit{like}\}\\
 & \texttt{passive\_1} & None of \{\textit{communicat}, \textit{suffer}, \textit{compet}, \textit{shout}, \textit{laugh},\\
& \texttt{passive\_2} &\ \textit{scream}, \textit{complain}, \textit{compromis}, \textit{grin}, \textit{chat}\} in sentence\\
\texttt{binding} & \texttt{principle\_A\_case\_1} & *Is the shorter of the two sentences\\
& \texttt{principle\_A\_case\_2} & *Is the longer of the two sentences\\
 & \texttt{principle\_A\_} & (Last word ends in \textit{s} and (1st word is any of \texttt{pl\_det}\\
 & \texttt{c\_command} & or the 2nd word is \textit{lot})) or 2nd to last word ends in \textit{s})\\
& \texttt{principle\_A\_domain\_1} & *Is the shorter of the two sentences\\
 & \texttt{principle\_A\_domain\_2} & *Is the shorter of the two sentences\\
& \texttt{principle\_A\_domain\_3} & Does not contain \textit{that}\\
 & \texttt{principle\_A\_reconstruction} &  4th word does not end in \textit{ed} nor \textit{'t}\\
\texttt{control\_raising} & \texttt{a\_obj\_raising} & Does not contain one of \texttt{verb\_set}\\
& \texttt{a\_subj} & Contains one of \texttt{subj\_words} or \{\textit{appear}, \textit{sure},\\ 
 \texttt{(existential\_}  & \texttt{subj\_raising} & \textit{threaten}, \textit{look}\}\\
 \texttt{there = a)} & \texttt{expletive\_it\_object\_raising} & Does not contain one of \texttt{verb\_set}\\
& \texttt{tough\_vs\_raising\_1} & Does not contain one of \texttt{subj\_words}, nor \textit{apt}\\
& \texttt{tough\_vs\_raising\_2} &  Contains one of \texttt{subj\_words}, or \textit{apt}\\
\texttt{determiner\_noun\_} 
& \texttt{a\_irregular\_1} & Does not end in \textit{that} followed by (one of \\
\texttt{agreement = b} & \texttt{b\_irregular\_2} & \{\textit{people}, \textit{women}, \textit{men}, \textit{children}\} or a word ending \\
 &\texttt{b\_with\_adj\_irregular\_1} & in \textit{ses}) nor in \{\textit{those}, \textit{these}\} followed by (a word \\ 
 & \texttt{b\_with\_adj\_irregular\_2} & ending in \textit{is} or not with \textit{s} at all)\\
& \texttt{b\_with\_adjective\_1} & Does not end in \textit{that} followed by a word ending in\\
& \texttt{b\_with\_adj\_2} & a letter other than \textit{i} followed by \textit{s} nor in\\
& \texttt{b\_1} & \{\textit{those}, \textit{these}\} followed by (a word ending in\\
& \texttt{b\_2} & \textit{is} or not with \textit{s} at all)\\
\texttt{ellipsis} & \texttt{ellipsis\_n\_bar\_1} & Last word in \texttt{num\_quant}\\
& \texttt{ellipsis\_n\_bar\_2} & Last word has already occurred in sentence\\
\texttt{filler\_gap\_} & \texttt{wh\_questions\_object\_gap} & Does not contain \textit{wh}\\
\texttt{dependency} & \texttt{wh\_questions\_subject\_gap} & Does not contain \textit{wh}\\
& \texttt{wh\_vs\_that\_no\_gap = c} & Does not contain \textit{wh}\\
& \texttt{c\_long\_distance} & Does not contain \textit{wh}\\
& \texttt{wh\_vs\_that\_with\_gap = d} & Contains \textit{wh}\\
& \texttt{d\_long\_distance} & Contains \textit{wh}\\
\texttt{irregular\_forms} & \texttt{irregular\_past\_} & If 1st word is \textit{the}, then 2nd word ends in \textit{n}, \\
& \texttt{part\_adj} & otherwise 2nd word must not end in \textit{n} \\ 
& \texttt{irregular\_past\_part\_verbs} & *Is the shorter of the two sentences\\
\texttt{island\_effects} & \texttt{adjunct\_island} & Last word is not \textit{about} and does not end in \textit{ing}\\
 & \texttt{e\_complex\_left\_branch} & 2nd word is not in \texttt{mod\_aux}\\
\texttt{(coordinate\_structure\_} & \texttt{e\_object\_extraction} & 2nd to last word is not \textit{and}\\
\texttt{constraint = e)} & \texttt{f\_echo\_question} & Does not start with \textit{Wh}\\
\texttt{(left\_branch\_} & \texttt{f\_simple\_question} & 2nd word is not in \texttt{mod\_aux}\\
\texttt{island = f)} & \texttt{sentential\_subject\_island} & Ends in \textit{ing} or \textit{ed} or \textit{with}\\
 & \texttt{wh\_island} & \textit{wh}, capitalized or not, occurs twice in sentence\\
\texttt{npi\_licensing} & \texttt{matrix\_question\_g} & 1st word in \texttt{mod\_aux}\\
 & \texttt{npi\_present\_1} & Does not contain the word \textit{ever}\\
\texttt{(npi\_licensor\_}& \texttt{npi\_present\_2} &  Does not contain the word \textit{ever} \\
\texttt{present = g)} & \texttt{only\_g} & 1st word is \textit{only}\\  
\texttt{(npi\_scope = h)} & \texttt{only\_h} & 1st word is \textit{only}\\
& \texttt{sentential\_negation\_g} & Does not contain the word \textit{ever}\\
& \texttt{sentential\_negation\_h} & Does not contain the word \textit{ever}\\
\texttt{quantifiers} & \texttt{a\_quantifiers\_1} & Does not contain \{\textit{each}, \textit{most}, \textit{all}, \textit{every}\} while \\
& \texttt{a\_quantifiers\_2} & also containing \{\textit{one}-\textit{ten}\} \\
& \texttt{superlative\_quantifiers\_1} & *Is the longer of the two sentences\\
& \texttt{superlative\_quantifiers\_2} & 1st word is not \textit{no}\\
\texttt{s-selection} & \texttt{animate\_subject\_passive} & Contains one of \texttt{people\_groups}\\
& \texttt{animate\_subject\_trans} & *Is the shorter of the two sentences\\

\texttt{subject\_verb\_agreement} & \texttt{i\_relational\_noun} & *Is the longer of the two sentences\\
 & \texttt{i\_relative\_clause} & The number of words that ends in \textit{s} is odd\\
\texttt{(distractor\_} & \texttt{irregular\_j\_1} & Contains no word ending in a letter other than \textit{i} and \\
\texttt{agreement = i)} & & followed by \textit{s} that is followed by a word ending in \textit{s} \\ 
\texttt{(plural\_subject\_} & \texttt{irregular\_j\_2} & None of \{\textit{people}, \textit{women}, \textit{men}, \textit{children}\} is\\
\texttt{verb\_agreement = j)} & & followed by a word ending in \textit{s} \\ 
 & \texttt{regular\_j\_1} & *Is the shorter of the two sentences \\
& \texttt{regular\_j\_2} & The number of words that ends in \textit{s} is odd \\
    \end{tabular}
    \caption{Linear Rule descriptions for BLiMP. Rules that require sentences to be compared are marked with an asterisk. Rules sometimes span across multiple rows. If one paradigm name is split across these rows, then the rule only corresponds to that paradigm. Otherwise the rule corresponds to all the paradigms listed in these rows. All variables (e.g. \texttt{verb\_set}, \texttt{subj\_words}) are defined in table \ref{tab:word_sets}.}
    \label{tab:blimp_rules}
\end{table*}

\begin{table*}[]
    \centering
    \small
    \begin{tabular}{l|l}
    \hline
Name & Content \\
    \hline
\texttt{verb\_set} & \{\textit{ask}, \textit{press}, \textit{entic}, \textit{prod}, \textit{obligat}, \textit{convinc}, \textit{badger}, \textit{compel}, \textit{sway}, \textit{order}\}\\ 
\texttt{subj\_words} & \{\textit{certain}, \textit{soon}, \textit{likely}, \textit{unlikely}, \textit{bound}, \textit{about}\} \\
\texttt{num\_quant} & \{\textit{one}-\textit{ten}, \textit{many}, \textit{few}, \textit{several}, \textit{more}, \textit{some}, \textit{lot}, \textit{fewer}\} \\
\texttt{mod\_aux} & \{\textit{had}, \textit{should}, \textit{is}, \textit{was}, \textit{can}, \textit{has}, \textit{will}, \textit{would}, \textit{could}, \textit{do}, \textit{does}, \textit{might}, \textit{were}, \textit{did}\} \\
\texttt{people\_groups} & \{\textit{men}, \textit{woman}, \textit{children}, \textit{teacher}, \textit{lad}, \textit{offspring}, \textit{student}, \textit{customer}, \textit{girl}, \textit{boy}\}
    \end{tabular}
    \caption{The sets of words represented by the variables used in table \ref{tab:blimp_rules}}
    \label{tab:word_sets}
\end{table*}

\newpage

\end{document}